# Addressing Overfitting on Pointcloud Classification using Atrous XCRF

Hasan Asy'ari Arief[a,*], Ulf Geir Indahl[a], Geir-Harald Strand[a,b], Håvard Tveite[a]

[a] *Faculty of Science and Technology, Norwegian University of Life Sciences, Ås, 1432, Norway; ulf.indahl@nmbu.no (U.G.I.);geir.harald.strand@nibio.no (G.H.S.); havard.tveite@nmbu.no (H.T.)*

[b] *Division of Survey and Statistics, Norwegian Institute of Bioeconomy Research, Ås, 1432, Norway*

[*] *Correspondence: hasanari@nmbu.no; Tel: +47-453-91-706*

**Abstract**

Advances in techniques for automated classification of pointcloud data introduce great opportunities for many new and existing applications. However, with a limited number of labeled points, automated classification by a machine learning model is prone to overfitting and poor generalization. The present paper addresses this problem by inducing controlled noise (on a trained model) generated by invoking conditional random field similarity penalties using nearby features. The method is called Atrous XCRF and works by forcing a trained model to respect the similarity penalties provided by unlabeled data. In a benchmark study carried out using the ISPRS 3D labeling dataset, our technique achieves 84.97% in term of overall accuracy, and 71.05% in term of F1 score. The result is on par with the current best model for the benchmark dataset and has the highest value in term of F1 score.

## 1. Introduction

The increased availability of high precision pointcloud data, including airborne LiDAR (light detection and ranging) data, has opened interesting possibilities for many applications such as generating digital elevation models (Podobnikar & Vrečko 2012), creating land use and land cover maps (Arief et al. 2018), 3D building reconstruction (Vosselman & Dijkman 2001), and scene understanding in large dynamic environments (Zhao et al. 2010)in.

Improving the visual quality and accuracy of automated pointcloud classification is an important topic in the computer vision, remote sensing and photogrammetry research communities. Many methods have been proposed to address this issue. Interesting examples includes SVM (Support Vector Machine) based pointcloud classification for urban areas (Mallet et al. 2008), the Random Forest combined with CRF (Conditional Random Field) approach (Niemeyer et al. 2014) for building detection, and the CNN (Convolutional Neural Network) approach for 3D semantic labeling (Yang et al. 2017), (Yousefhussien et al. 2017), (Zhao et al. 2018).

The use of DNNs (Deep Neural Networks) for pointcloud classification has attracted considerable attention in the last couple of years because of its potential for improving the quality of automated classification. The quantitative results for several pointcloud classification benchmark datasets, such as 3D Shapenets (Wu et al. 2015), ScanNet (Dai et al. 2017), S3DIS (Armeni et al. 2016), and ISPRS Vaihingen 3D Labeling (Niemeyer et al. 2014) show that the majority of high performance classifiers for all these datasets are based on some choice of DNN model.

An easy and yet robust implementations of DNN for pointcloud data is known as the PointCNN (Li et al. 2018). The PointCNN uses a so-called *X*-transformation to allow a convolution operator to work directly on the pointcloud data. In contrast to other methods such as voxel-based methods (Maturana & Scherer 2015) and raster based methods (Zhao et al. 2018), the PointCNN reduces significantly both the required amount of time for preprocessing and the memory usage. Such advantages are important for real-time classification of pointcloud data.

One of the main challenges of PointCNN and other DNN based models is that when only datasets of relatively limited size are available, they are highly vulnerable to overfitting. This is because such models usually include several million parameters, and robustly fitting such an amount of parameters requires a large number of training data. With our proposed Atrous XCRF method we overcome the overfitting problem by inducing controlled noise when training a DNN based classifier. The method works by retraining a validated model using unlabeled test data. The training supervision is directed by utilizing the hierarchical structure of the CRF penalty procedure (Krähenbühl & Koltun 2011). In our experiment with the ISPRS 3D labeling benchmark dataset, we get an Overall Accuracy (OA) of 84.91% and an F1-Score of 71.05% (Wolf & Jolion 2006).

The present paper is organized as follows: In section 2, we provide a brief review of DNN, PointCNN, and CRF modelling. We also explain the XCRF and our proposed Atrous XCRF method for handling the overfitting problem. In the following section, we describe the experiment, including the data source, preprocessing procedure, training strategies, and results analysis. Thereafter, we discuss the limitations and characteristics of our proposed method. Finally, we provide the conclusions and indicate potential improvements of our novel technique. The trained model and reproducible code are available at https://github.com/hasanari/A-XCRF.

## 2. Methodology

### 2.1. Brief review of deep neural networks

Deep neural networks (DNNs) or Deep Learning is an extension of a classical two-layer neural network (Blum & Rivest 1989) using more (and wider) layers with some important enhancements. In contrast to the classical networks, the architechture of DNNs typically includes up to thousands of layers and up to millions of parameters which are normally trained using the complete architecture in an end-to-end fashion to achieve the best possible classification performance (LeCun et al. 2015).

The DNNs that are currently the most popular for image classification tasks are Convolutional Neural Networks (CNN) (LeCun et al. 1995). The CNNs extend the classical neural network principles into an extraordinary powerful classifier (Krizhevsky et al. 2012). They involve three basic operations, namely, convolutions, pooling operations, and non-linear activation functions (Schmidhuber 2015).

A convolution operation is essentially a collection of dot product operations which allow a number of parameters (organized as a set of corresponding kernels) to aggregate on top of the feature maps provided by the previous layer to create the input to the subsequent layer. The convolution operations make it possible for a CNN to capture spatial autocorrelation phenomena in the data into the resulting CNN model.

The pooling steps in CNN modelling is necessary for reducing the spatial size of the input feature map. Taking the maximum value of the features to represent the derived combined features is usually referred to as "max-pooling". In addition to reducing the spatial size of the feature map (e.g. reducing the need for memory storage and computation), the pooling operations also create a spatially generalised representation of the data.

Another important aspect of CNN modelling is the non-linear activation functions associated with the computational nodes in the network. Their purpose is to obtain non-linearity in the transformations between the subsequent layers of the network. If omitted, a CNN (in fact any Neural Network) could be collapsed into a single linear transformation incapable of modelling the massively nonlinear phenomena present in most practical applications (Minsky & Papert 1987). Most applications of CNNs use (some version) of a Rectified Linear Unit (ReLU) (Dahl et al. 2013) as their non-linear activation functions.

A CNN model is trained using gradient descent (Recht et al. 2011), which calculates the contribution of weight kernels towards the final loss value. The parameter update of a CNN model is based on the chain rule, using the backpropagation algorithm (LeCun et al. 2015). The cross-entropy loss function (De Boer et al. 2005) is used to measure the precision of the trained CNN model, and reflects the degree of correspondence between the true and predicted class labels.

## 2.2. Brief review of the PointCNN

The convolution operation of CNNs is efficient for capturing spatial correlations from gridded datasets, such as images. The PointCNN introduces a modified version of the convolutions used in CNNs, which we will refer to as X-Conv. The X-Conv modifies the ordinary CNN convolution to work for irregular and unordered datasets such as pointcloud data (Li et al. 2018). Because the X-Conv can be used directly with irregular data, the need for preprocessing is significantly reduced.

| **ALGORITHM 1**: X-Conv Operator (a trainable feature extraction block included in the architecture), taken from the PointCNN paper (Li et al. 2018). | |
|---|---|
| **Input** : $K$, p, **P**, **F** | |
| **Output** : $\mathbf{F}_p$ | Features "projected", or "aggregated", into a representative point $p$. |
| 1: $\mathbf{P_o} \leftarrow \mathbf{P} - p$ | Move **P** to a local coordinate system with $p$ as origo. |
| 2: $\mathbf{F}_\delta \leftarrow MLP_\delta(\mathbf{P_o})$ | **Individually** lift each point into $C_\delta$ dimensional space. |
| 3: $\mathbf{F}_* \leftarrow [\mathbf{F}_\delta, \mathbf{F}]$ | Concatenate $\mathbf{F}_\delta$ and **F**, $\mathbf{F}_*$ is a K × ($C_\delta + C_1$) matrix. |
| 4: $\mathbf{X} \leftarrow MLP(\mathbf{P_o})$ | Learn the K × K, **X**-transformation matrix. |
| 5: $\mathbf{F}_X \leftarrow \mathbf{X} \times \mathbf{F}_*$ | Weigh and permute $\mathbf{F}_*$ with the learnt **X**. |
| 6: $\mathbf{F}_p \leftarrow Conv(\mathbf{K}, \mathbf{F}_X)$ | Finally, typical convolution between **K** and $\mathbf{F}_X$. |

Similar to a CNN convolution, the X-Conv includes the calculation of inner products

(element-wise product and summation operations) between feature maps and the convolution kernels. The X-Conv takes into consideration neighbouring points among the features of interests, and finally transforms these features by a Multi-Layer Perceptrons (MLP) (Atkinson & Tatnall 1997).

The X-Conv operation is described in Algorithm 1. Here *p* denotes the input point, **P** denotes the *K* neighbor points, **K** denotes the trainable convolution kernels and **F** denotes the previous feature representations of the *K* neighboring points.

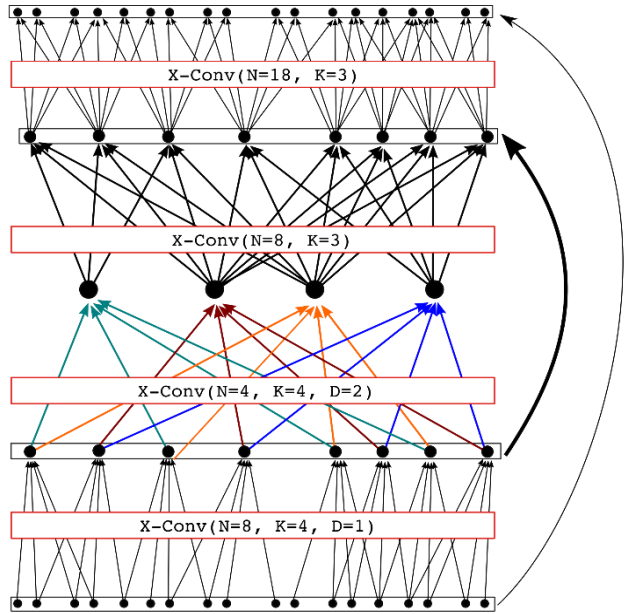

**Figure 1.** PointCNN architecture for point level prediction.

The PointCNN for pointcloud segmentation is organized as a stack of several Conv-DeConv blocks (Noh et al. 2015) using the U-Net architectural design (Ronneberger et al. 2015). Similar to the U-Net, the Conv blocks are used to generate the global feature maps by maintaining local connectivity, while the DeConv blocks are used to propagate the global features into point level predictions. For the PointCNN, both the Conv and the DeConv blocks involve the X-Conv operation, but with a different number of points and receptive fields. Similar to the U-Net design, the output from the previous Conv block is forwarded not only to the next Conv block but also to the corresponding DeConv block, see Fig. 1 for details (*K* denotes the number of nearby points, *N* denotes the number of output points, and *D* denotes the atrous distance). It should be noted that the PointCNN also includes the dropout technique (Srivastava et al. 2014) for its Fully Connected (FC) layers as a regularization module to improve the accuracy of the resulting classifier.

2.3. *A brief review of the Fully Connected CRF*

A Conditional Random Field (CRF) is a probabilistic graphical model often used for sequence segmentation and labeling, capable of relaxing the strong independence assumptions of a graph model (Lafferty et al. 2001). A fully connected CRF (Krähenbühl & Koltun 2011) is a variant of CRF that is applied for the fully connected graph. For example, if the fully connected CRF is implemented on an image with its probability maps, the conditional penalty for a pixel in the image is calculated from the similarities and distances between that pixel and all the other pixels of the image.

A fully connected CRF is represented using a Gibbs distribution and defined as:

$$E(x) = \sum_i \psi_u(x_i) + \sum_{i<j} \psi_p(x_i, x_j),$$

where $E(x)$ denotes the Gibbs energy, $x$ denotes the label assignment (of the entire image), $i$ and $j$ denote pixel locations, $\psi_u(x_i)$ denotes the unary potential on pixel $i$ (the unary potential is the result of an independent classifier) and $\psi_p(x_i, x_j)$ denotes the pairwise potential between the labels $x_i$ and $x_j$ and is defined as:

$$\psi_p(x_i, x_j) = \mu(x_i, x_j) \sum_{m=1}^{K} w^{(m)} k^{(m)}(f_i, f_j),$$

$k^{(m)}(f_i, f_j)$ is a Gaussian filter (Paris & Durand 2006) that calculates the similarity between the feature vectors $f_i$ and $f_j$ for the pixels $i$ and $j$, respectively. For multi-class classification, the Gaussian filter is implemented as contrast sensitive two-kernel potentials using a weighted ($w^{(m)}$) Gaussian filter (Tomasi & Manduchi 1998). The pairwise potential is also weighted by the compatibility function denoted as $\mu(x_i, x_j)$ using a Potts model (Krähenbühl & Koltun 2011). $\mu(x_i, x_j)$ penalizes nearby similar pixels that have different labels. A fully connected CRF is trained using iteratives Mean Field Approximation, and the Gaussian filters are computed using the permutahedral lattice (Adams et al. 2010), a high dimensional filtering algorithm. The details are explained in the above mentioned paper describing the fully connected CRF (Krähenbühl & Koltun 2011).

A variant of a fully connected CRF, implemented by using convolution operations and trained end to end using stack of convolution layers, is the so-called CRF Recurrent Neural Network (CRF-RNN) (Zheng et al. 2015). The iterative CRF mean field operation of the CRF-RNN is structured as a stack of CNN layers, and the Gaussian filter is implemented using the permutahedral lattice where the filter coefficients convolves the weighted values on the lattice space. The mean-field iteration takes a weighted sum of the previous outputs for each class label, corresponding to a 1 × 1 convolution on every class label. The compatibility transform can also be seen as the convolution of a Potts model with the outputs calculated in the previous step. Finally, the update operation of a unary potential is obtained by adding the pairwise potentials and the current unary potential together. The updated outputs are then used as the new unary potentials. The described operations are organized into a Recurrent Neural Network (RNN) architecture (Mikolov et al. 2010), so that gradient descent can be used to update the weighted Gaussian filter and the CRF compatibility matrix.

*2.4. Training the artificial labels using Atrous XCRF*

The Atrous XCRF (A-XCRF) can be explained as a variant of CRF-RNN, which has the same properties of calculating the pairwise similarities and penalizing according to the predictions. The main difference between the two is that the A-XCRF does not require a permutahedral lattice. The pairwise penalty of A-XCRF is implemented using a hollow matrix and one-hot encoding of the predicted label, and the method is used to refine a trained DNN model.

The X term in XCRF is associated with the X-Transformation in the PointCNN which utilizes the nearby points to create features of interest, and uses atrous (Arief et al. 2018) indices for point selections. For point data, the atrous approach (see Fig. 2) means that the selected indices are not necessarily close to each other but closest by some number of intermediate (unselected) points, see (Arief et al. 2018) for a detailed explanation of atrous indices for raster / grid data.

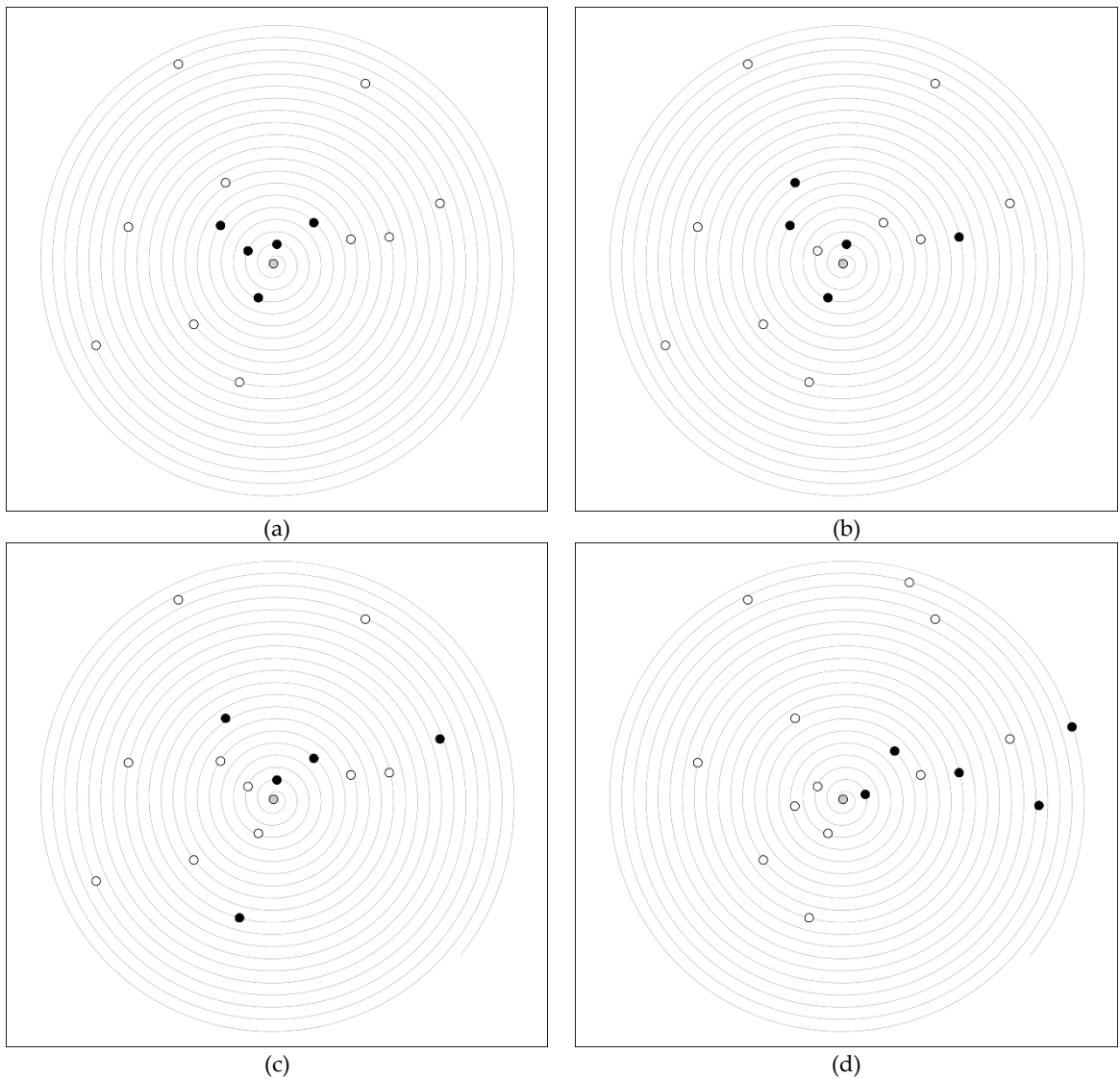

**Figure 2.** Atrous indices for point selections with *K*=5. The selected points are marked as black dots with the center point marked as grey, (a) *D*=1, (b) *D*=2, (c) *D*=3, and (d) *D*=4.

While CRF-RNN is fully connected, XCRF is not fully connected and only considers the specified *K* number of nearby points. The intuition justifying its application is that a patch of pointcloud data can be spread out in a very large region, and the points being far apart should not influence each other very much. By ignoring such distant pairs of points, we gain substantial savings computationally and in memory consumption. The XCRF is outlined in Algorithm 2. **P, U, F** denote the matrices containing input points (**P**), current unaries potential (**U**), and existing features for each of the points (**F**), respectively. *K, D, r,* **I** denote the number of nearby points (*K*), the atrous distance between point indices (*D*), the number of update iterations (*r*), and nearby point indices (**I**), respectively.

| | |
|---|---|
| **ALGORITHM 2**: XCRF Algorithm | |
| **Input** :**P, U, F**, *K*, **I**, *r*, *D* | |
| **Output** :**U**$_1$ | Updated unary potentials respecting the **K** points' similarity |
| 1: **P**$_I$ = gather(**I**, *D*, *K*) | For a point p in **P**, gather the *KxD* nearest point indices **I**, sorted on increasing distance, skipping *D* points for each gathered point index (**I**$_g$). |
| 2: **D**$_C$ ← distance(**P, F, P**$_I$) | Calculate the euclidean distances between **p** and **P**[**I**$_g$], and the distances between the feature values of **F**$_p$ and **F**[**I**$_g$]. |
| 3: **B**$_f$, **S**$_f$ ← gaussian(**D**$_C$) | Implement the Gaussian filtering (Krähenbühl & Koltun 2011) on **D**$_C$, with **B**$_f$ for the bilateral filter and **S**$_f$ for the spatial filter. |
| 4: **G**$_w$ ← **B**$_f$ × **W**$_b$ + **S**$_f$ × **W**$_s$ | Passing the gaussian weights (**W**$_b$ and **W**$_s$) on the previous outputs, as weighted Gaussian (**G**$_w$). |
| 5: **U**$_1$ ← **U** | Duplicate original Unary (**U**). |
| 6: Update iteration as RNN, range(*r*): | |
| 6.1: **U**$_s$ ← softmax(**U**$_1$) | Normalize the unary potential with the softmax function (LeCun et al. 2015). |
| 6.2: **W**$_u$ ← OneHot(**U**$_s$) * **W**$_c$ | Calculate hollow weighted unaries by using the dot product of the one hot encoding of the **U**$_s$ and a hollow weighted matrix (**W**$_c$). |
| 6:3 **U**$_G$ ← **U**$_s$ × **G**$_w$ | Pass the normalized unary to the weighted Gaussian output. |
| 6:4 **U**$_p$ ← **U**$_G$ * **W**$_u$ | Calculate the pairwise penalty as a dot product of the weighted Gaussian and the compatibility hollow matrix. |
| 6:5 **U**$_1$ ← **U** - **U**$_p$ | Update the unary values with the pairwise penalty, and after *r* iterations, return **U**$_1$ as the new unaries. |

In line 1, for point p in **P**, **P**$_I$ gathers the indices of *K* nearby points according to the specified atrous distance (*D*) from the list of indices of the *KxD* nearest points (sorted on the distance from p), as illustrated in figure 2. In line 2-3, the similarity penalties between a point and its *K* neighbours is calculated using the Gaussian bilateral and spatial filters denoted $B_f$ and $S_f$, respectively. These filter are defined as:

$$B_f = \exp\left(-\frac{|p_i - p_j|^2}{2\theta_\alpha^2} - \frac{|I_i - I_j|^2}{2\theta_\beta^2}\right), \text{ and}$$

$$S_f = \exp\left(-\frac{|p_i - p_j|^2}{2\theta_\gamma^2}\right)$$

where, $p_i$ denotes the spatial (x, y, z) coordinates of point $p_i$ and $I_i$ denotes the feature vector of $p_i$. $\theta_\alpha$, $\theta_\beta$, and $\theta_\gamma$, are the normalizing constants for the euclidean distances. $B_f$ and $S_f$ act as similarity penalties because their values increase as the associated euclidean similarities decrease (dissimilar features) and decrease as the euclidean similarities increase. In other words, the penalties are larger for nearby and similar points and smaller for more remote or dissimilar points. In step 4 of the XCRF-algorithm, the Gaussian outputs are weighted with kernels (**W**$_b$ and **W**$_s$) being updated in the training process of the complete architecture. In step 5, the original unaries are duplicated (to be updated in step 6). In steps 6, the unary potentials and the similarity penalties are combined to update the pairwise unary potentials as a RNN iteration, see also (Zheng et al. 2015). The effect of step 6 is that the original unary potential are recursively updated using the weighted gaussian filters and the similarity label penalties using the hollow compatibility matrix.

The Gaussian filters in the XCRF algorithm sharpens the edge between two dissimilar points based on their normalized euclidean similarity distance. The weighting coefficients determine the amount of penalization according to the similarities. The hollow matrix and the one hot encoding output, on the other hand, work by penalizing label differences with weighted penalties and do not penalize equal labels, similar to the Potts model of the fully connected CRF (Krähenbühl & Koltun 2011). XCRF can therefore be seen as a strict penalty procedure that is particularly sensitive to nearby similar points having different labels.

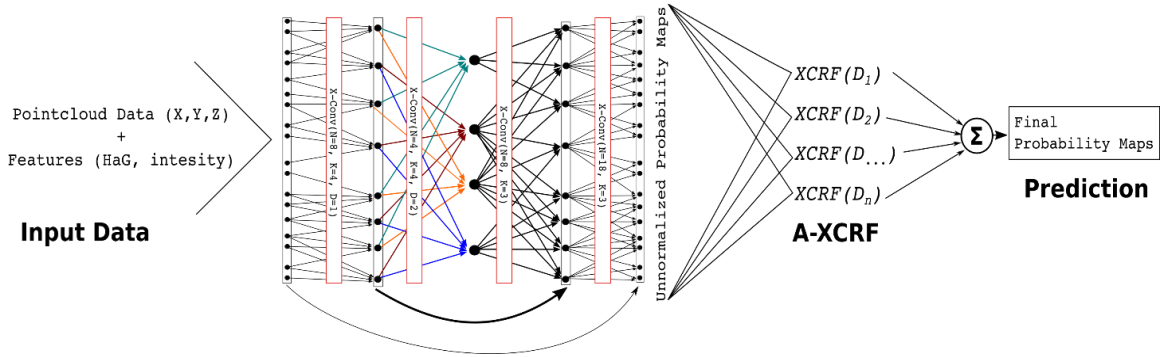

**Figure 3.** Implementation design of a full A-XCRF architecture.

The proposed A-XCRF block builds on the XCRF algorithm, but includes a modification to make it work as a refinement block for DNN architectures, similar to the CRF-RNN. The main difference is that, in addition to the recurrent structure, the Atrous XCRF requires multiple *D*s (atrous distances between point indices) to implement the hierarchical structure of the XCRF, see Fig. 3

With $U_{\text{final}}$ denoting the final unary values, $n$ denoting the number of different *D*'s in *D*, and *Ps* denoting the collection of XCRF parameters as described for Algorithm 2, the A-XCRF is simply defined as:

$$U_{\text{final}} = \sum_{i=1}^{n} \text{XCRF}(Ps, D_i).$$

A two-step process is required to train the A-XCRF. The first step is seeking the best possible (in the validated sense) model for a DNN architecture by training the model with a split-validation approach. Inclusion of the split-validation part is important to prevent against harmful overfitting.

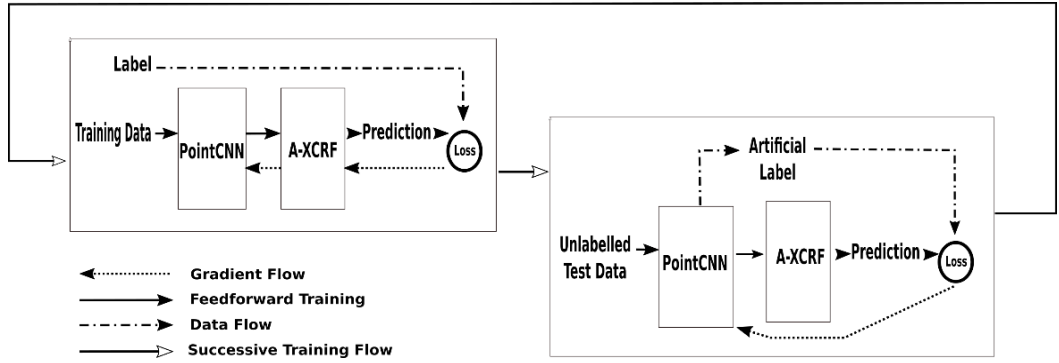

**Figure 4.** Training flow for the PointCNN with A-XCRF using labeled and unlabeled data.

The second step includes training of the XCRF parameters against the validated model obtained in the first step. The training is carried out using the unseen data (unlabeled test data). The labels for these data are the predicted labels of the validated model, called artificial labels, see Fig. 4. The quality of the artificial labels obviously depends on the accuracy of the validated model from the first step.

The underlying idea of the second step is to introduce noise in the validated model by invoking the strict pairwise penalties of XCRF on the unseen data. The validated model is therefore forced to respect the XCRF penalties when fitted to the training data. By using this approach, one can invoke the differences between the calculated DNN probabilities obtained before and after they have passed through the A-XCRF refinement block.

Invoking the XCRF as a controlled noise generator can be useful when training a DNN model. This is particularly the case for PointCNN architechtures where features are generated from nearby points, limiting the potential for data augmentations (rotations and scalings do not affect neighbour relationships). The introduction of a controlled amount of noise acts as a regularizer on the DNN model that helps to overcome the ordinary limitations of pointcloud data augmentation.

In order to maintain the accuracy of a validated model, the training process in the second step successively swaps between the two datasets (the training data and the unseen data with its artificial labels). The cross entropy loss function is used for training with both datasets. The parameter update using the backpropagation algorithm, on the other hand, works differently for the two datasets. For the training data, the backpropogation updates both the DNN kernels and the XCRF kernels. However, for the unlabeled data, the algorithm only updates the DNN kernels while the XCRF kernels are kept fixed, see Fig 4. By this approach, the updating of the DNN kernels works well with the training data, and at the same time respects the pairwise penalties of XCRF associated with the unseen data.

The trained model, including both the DNN- and A-XCRF parts of the architecture, makes up the resulting point classifier. The classifier works by passing the unary outputs from the DNN to the A-XCRF block for calculating the final unary potentials or the non-normalized prediction maps. A softmax normalization of the prediction maps generates the final class probability maps and the index of the highest class probability identifies the predicted class label for a particular point in the pointcloud.

# 3. Experiments and Results

## 3.1. Benchmark dataset

The proposed method was evaluated using an open benchmark dataset, the ISPRS 3D labeling dataset (Cramer 2010). The labeling was provided by Niemeyer (Niemeyer et al. 2014). The dataset is an Airborne Laser Scanning (ALS) dataset acquired using a Leica ALS50 system with mean flying height 500 m above Vaihingen village in Germany. The dataset has a median point density of 6.7 points/m² and has nine classes for the 3D labeling task (powerline, low vegetation, impervious surfaces, car, fence/hedge, roof, facade, shrub, and tree). The dataset was divided into two parts, the training data and the test data with 753,876 points and 411,722 points, respectively. Both parts contained spatial coordinates (XYZ), intensity values and the number of returns for each point. Table 1 shows the class distribution for the training data.

| Class | Number of Points | |
|---|---|---|
| | Training Data | Test Data |
| Powerline | 546 | - |
| Low Vegetation | 180,850 | - |
| Impervious Surfaces | 193,723 | - |
| Car | 4,614 | - |
| Fence/Hedge | 12,070 | - |
| Roof | 152,045 | - |
| Façade | 27,250 | - |
| Shrub | 47,605 | - |
| Tree | 135,173 | - |
| Total | 753,876 | 411,722 |

**Table 1.** Class distribution of the Vaihingen 3D labeling dataset.

The benchmark dataset was evaluated using the Overal Accuracy (OA) and the F1 score. The OA is the percentage of points that were correctly classified while the F1 score is the harmonic average of precision and recall. The F1 score is more sensitive to the unbalanced class distribution, as observed in the training data.

The F1 score is defined as follows:

$$precision = \frac{TP}{TP+FP},$$

$$recall = \frac{TP}{TP+FN}, \text{ and}$$

$$\text{F1 Score} = 2 * \frac{precision * recall}{precision + recall}$$

where TP, FP and FN denote True Positive, False Negative and False Positive, respectively.

## 3.2. PointCNN and XCRF parameter setting

PointCNN works by training a batch of point blocks at once. It is therefore necessary to slice the dataset into point blocks before the model can be trained. For this purpose, 25m by 25m splitting blocks were used. The choice of block size was based on the point density of the dataset and the fact that PointCNN resamples and trains 2048 points in one forward pass (training iteration).

The training data was first sliced using 100m by 100m blocks (ignoring the Z axis), resulting in 12 blocks, each block containing between 25 000 and 120 000 points. Of these blocks, 80% were used for training and the remaining for validation, ensuring that there is no overlap between the training and validation data. Each of the 100m by 100m blocks were then sliced using 25m by 25m splitting blocks. Re-slicing were also performed by moving the initial slicing block by 12.5m (half sliced block) overlapping all the edges of the previous sliced blocks, hence increase the number of blocks and data points by repetition. The slicing process produced 286 and 44 blocks for the training and validation data, respectively. The number of points per 25m by 25m block varied between 1300 and 9000 points. The test data was also sliced using the same process, and produced 119 blocks of test data. It should be noted that the spatial coordinates for the points were transformed to local coordinate systems with origin at the center of the block the point belongs to.

For every training batch, 2048 points per block were randomly sampled by PointCNN. For blocks that have less than 2048 points, points were resampled with replacement. One 11GB GeForce GTX 1080 Ti graphics card was used for training, and the batch size was set to only six, due to capacity limitations of the GPU. It should be noted that increasing the batch size could improve the results, because PointCNN uses Batch Normalization (Ioffe & Szegedy 2015) to reduce the internal covariate shift of the DNN parameters, and the technique performs better for a bigger batch size. Unless otherwise mentioned, the Tensorflow (Abadi et al. 2016) DNN library was used, and the training process started with a learning rate of 0.005, reducing it by 20% for every 5000 iterations, with a minimum learning rate of $1e^{-6}$.

The XCRF weight parameters were initialized to one, assuming an equal contribution from each parameter toward the final loss. The compatibility matrix was also initialized to one, but multiplied with a zero diagonal matrix to produce a hollow matrix. The Gaussian filter parameters ($\theta_\alpha$, $\theta_\beta$, and $\theta_\gamma$) were initialized using a grid search, see (Krähenbühl & Koltun 2011) for Gaussian filter initialization strategies. It should be noted that the filter parameters decide the size of the areas to be considered when calculating the pairwise similarities.

The XCRF uses five iterations ($r$) to update the unary potentials both for the XCRF parameters update and for generating the final predictions. The A-XCRF is implemented on six hierarchies, with $D$ equal to 1, 2, 3, 4, 8 and 16, respectively. For all of the levels a $K$ (number of neighboring points) value of 64 was used, therefore the farthest neighboring point that is considered for the sixth level of the A-XCRF is the 1024th point ($K$=64 by $D$=16).

### 3.3. Training Strategies and Results

In order to test our proposed method, we trained the original PointCNN and the PointCNN with A-XCRF separately and analyzed the results to quantify the improvements. For the remainder of this paper, we use PointCNN for the original PointCNN and A-XCRF for the PointCNN with Atrous XCRF. Both techniques were trained using x, y, z and two additional features: 1) Height Above Ground (HaG) generated using TerraScan (Terrasolid 2016), and 2) intensity. Both of these additional features were normalized to the range [-0.5, 0.5].

PointCNN and A-XCRF were trained using the split-validation dataset and the training was concluded after ten consecutive training epochs without any change in the validation accuracies. We also trained the PointCNN with CRF-RNN using the same dataset and stop procedure. The PointCNN with CRF-RNN was trained end-to-end, as recommended in the CRF-RNN paper (Zheng et al. 2015).

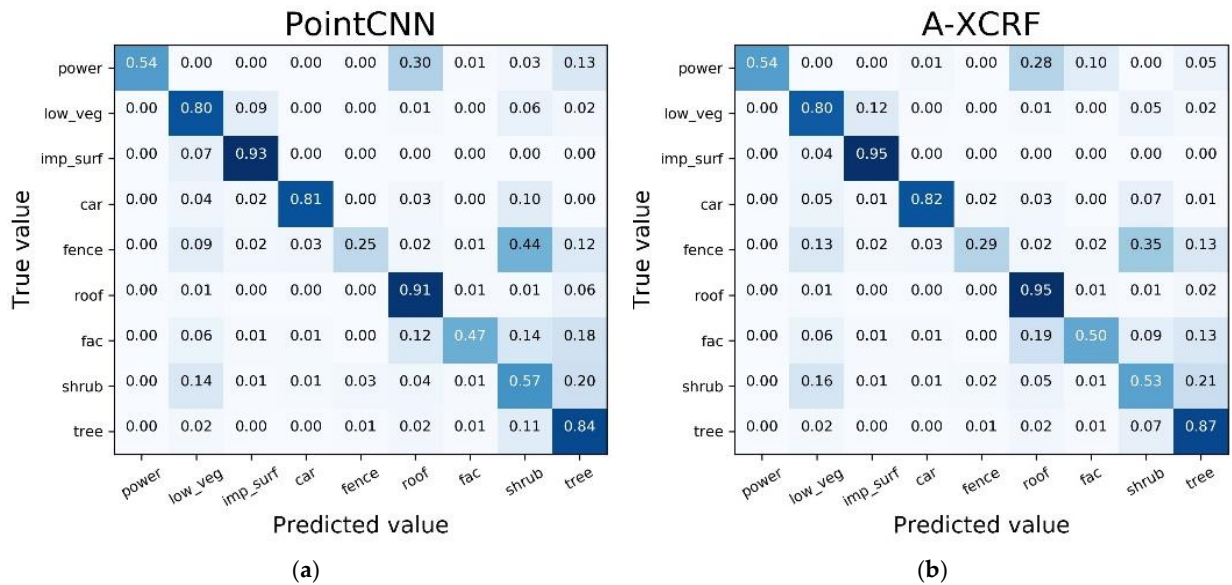

**Figure 5.** Comparison of the confusion matrices of **(a)** PointCNN and **(b)** A-XCRF.

| Class | PointCNN | CRF-RNN | A-XCRF |
|---|---|---|---|
| Powerline | 61.45 | **68.46** | 62.97 |
| Low Vegetation | **82.71** | 81.70 | 82.59 |
| Impervious Surfaces | 91.84 | **92.08** | 91.91 |
| Car | **75.84** | 75.70 | 74.86 |
| Fence/Hedge | 35.90 | 38.16 | **39.87** |
| Roof | 92.68 | 93.45 | **94.48** |
| Façade | 57.83 | 58.75 | **59.33** |
| Shrub | 49.14 | 50.12 | **50.75** |
| Tree | 78.10 | 79.45 | **82.69** |
| Average F1 | 69.50 | 70.87 | **71.05** |
| OA | 83.33 | 83.59 | **84.97** |

**Table 2.** OA and F1 scores for PointCNN based techniques on the test data of the Vaihingen Dataset. All cells except the last two rows show F1 scores.

Table 2 shows a comparison of the results on the test data for A-XCRF, PointCNN and PointCNN with CRF-RNN (CRF-RNN in the table). Fig. 5 shows the confusion matrices of PointCNN and A-XCRF. It can be noticed that A-XCRF improves on PointCNN for the majority of the classes. Similar to the PointCNN, A-XCRF has a problem with classifying the Powerline points. This is because the atrous approach of both models create holes in the group of neighbouring points while gathering nearby points, and this does not work well for linear features such as powerlines. CRF-RNN calculates similarity using all the nearby points, which seems to be a better strategy for such features.

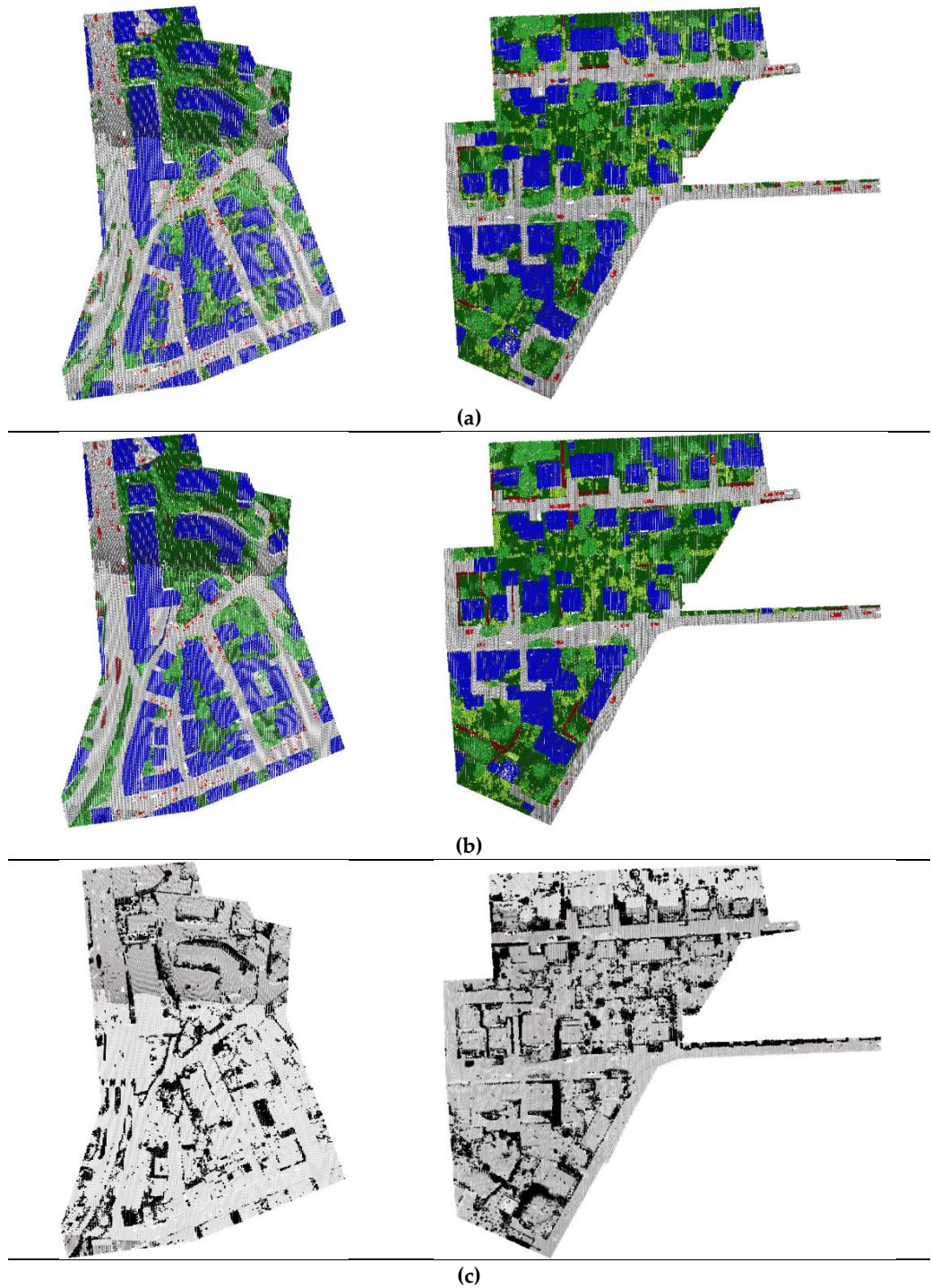

**Figure 6**. Classification map generated by **(a)** the A-XCRF method and **(b)** Vaihingen label test set, and **(c)** error map. Classes and colors: power line (orange), low vegetation (dark green), impervious surface (light gray), car (red), fence/hedge (dark red), roof (blue), façade (blue navy), shrub (green yellow), and tree (lime green).

Based on Table 2, it is clear that A-XCRF only offers a slight (between 1 and 2 percent) improvement in performance compared to the other techniques. However, A-XCRF is used for post-processing, and could be used as an extension to other machine learning-based classifiers, making it interesting as a tool for improving the quality of any classifiers. Fig. 6 shows the classification maps and error map produced by the A-XCRF.

*3.4. ISPRS and Benchmark Results*

| Class | ISS_7 | UM | HM_1 | LUH | RIT_1 | WhuY4 | A-XCRF |
|---|---|---|---|---|---|---|---|
| Powerline | 54.4 | 46.1 | **69.8** | 59.6 | 37.5 | 42.5 | 62.97 |
| Low Vegetation | 65.2 | 79.0 | 73.8 | 77.5 | 77.9 | **82.7** | 82.59 |
| Impervious Surfaces | 85.0 | 89.1 | 91.5 | 91.1 | 91.5 | 91.4 | **91.91** |
| Car | 57.9 | 47.7 | 58.2 | 73.1 | 73.4 | 74.7 | **74.86** |
| Fence/Hedge | 28.9 | 5.2 | 29.9 | 34.0 | 18.0 | **53.7** | 39.87 |
| Roof | 90.9 | 92.0 | 91.6 | 94.2 | 94.0 | 94.3 | **94.48** |
| Façade | - | 52.7 | 54.7 | 56.3 | 49.3 | 53.1 | **59.33** |
| Shrub | 39.5 | 40.9 | 47.8 | 46.6 | 45.9 | 47.9 | **50.75** |
| Tree | 75.6 | 77.9 | 80.2 | **83.1** | 82.5 | 82.8 | 82.69 |
| Average F1 | 55.27 | 58.96 | 66.39 | 68.39 | 63.33 | 69.2 | **71.05** |
| OA | 76.2 | 80.8 | 80.5 | 81.6 | 81.6 | 84.9 | **84.97** |

**Table 3.** A quantitative comparison between A-XCRF and other methods on the Vaihingen dataset. All cells except the last two rows show the per-class F1 score.

A quantitative comparison between our method (A-XCRF) and other methods tested on the Vaihingen dataset are shown in Table 3. ISS_7 (Ramiya et al. 2016) applied a supervoxel method on the point cloud data using the voxel cloud connectivity algorithm; then the local connectivity of the supervoxels were used to generate segmented objects; and finally, the generated segments were classified using machine learning techniques. UM (Horvat et al. 2016) used a combination of the pointcloud attributes, textural properties and geometrical attributes generated using morphological profiles, and trained the features using One-vs-One (OvO), a multiclass machine learning technique. HM_1 (Steinsiek et al. 2017) used k-nearest neighbors (KNN) to select neighboring points and eigenvalue-based features to generate geometrical features at the point level, then conducted the contextual classification using CRF. The LUH (Niemeyer et al. 2016) applied hierarchial higher order CRF by using two independent CRFs on the points and segments level (clustered points), respectively. It should be noted that the majority of the techniques listed in Table 3 did not explain their stop procedure or their use of validation data. This information would be helpful when trying to replicate their result, as well as when testing how A-XCRF would behave as a post-processing step for these methods.

RIT_1 (Yousefhussien et al. 2017) and WhuY4 (Yang et al. 2018), are DNN based techniques for pointcloud classification. RIT_1 used a 1D-fully convolutional network with terrain-normalized points and spectral data. WhuY4 used a multi-scale CNN on the point to raster representations, utilizing geometrical features such as planarity, sphericity, and variance of deviation angles, in addition to the HaG and intensity values.

It should be noted that there is another deep learning method called NANJ2 (Zhao et al. 2018) not shown in Table 3. The method was omitted from the table because it ignored the Powerline class (Zhao et al. 2018). The accuracy and the average F1 can therefore not be compared with the results from other methods. The NANJ2 generated deep features learned from the height, intensity, and roughness attributes and trained the features using multi-scale CNN.

## 4. Discussion

### 4.1. Limitations

The improvement offered by A-XCRF was about 2 %, which may not be considered as very significant in terms of accuracy improvement. However, the use of artificial labels and XCRF similarity penalties to re-train a validated model is a novel idea and provides a basis for a future research direction.

In modelling with neural network architechtures, parameter update and gradient descent works by tracking the gradient flows of the parameters and updating the parameters based on the accumulative loss caused by those parameters. Consequently, the distances between feature vectors are not emphasized in neural networks modeling. Invoking the similarity penalties between features by using the CRF technique as a post-processing procedure to improve classification accuracy is not a new idea (Krähenbühl & Koltun 2011), (Chen et al. 2016), (Zheng et al. 2015). However, this paper extends the idea by using unlabeled data with similarity penalties to improve the results, and the improvement is confirmed in Table 2 and 3.

Geometrical features calculated from the nearby $K$ points can be used to strengthen the point similarity penalties. Sphericity, planarity, and deviation angle variance, proposed in (Yang et al. 2018), are examples of geometrical features that can be generated and used as additional features for the XCRF.

Including the geometrical features when calculating the pairwise similarity penalties may help the XCRF to provide better classification results on classes that have a linear geometrical shape, such as Powerline, Façade, and Fence/Hedge. Figure 5 shows the unsatisfactory classification accuracies of those particular classes. Including geometrical features, such as planarity and sphericity, that are capable of detecting linearity from neighbouring points in the pointcloud could contribute to overcome this problem. However, more research is needed to define a well-suited similarity penalty formula for such geometrical features.

The Gaussian kernel defined in Eq. 3 in (Krähenbühl & Koltun 2011) includes a contrast sensitive two-kernel potential combining a bilateral and a spatial filter that gives higher penalties for smaller distances. This approach does not seem to be very well suited for stacking many different features with different characteristics, because stacking them in the mentioned kernel seems to reduce the impact on the final similarity penalties. In other words, adding many different features in the Gaussian kernel are likely to both increase dissimilarity values and reduce the pairwise penalties. More work is required to derive similarity formulas that works better with generated and dissimilar geometrical features.

### 4.2. End-to-end Atrous XCRF

When used as a post-processing module, the A-XCRF seems to be well suited to improve the prediction quality of machine learning based classifiers. However, the two loss functions of A-XCRF are making it more complicated to formulate the method as an end-to-end training process. This is because both loss functions update the same kernel (DNN parameters) and distributing the losses in an end-to-end fashion would results in intractable parameter updates.

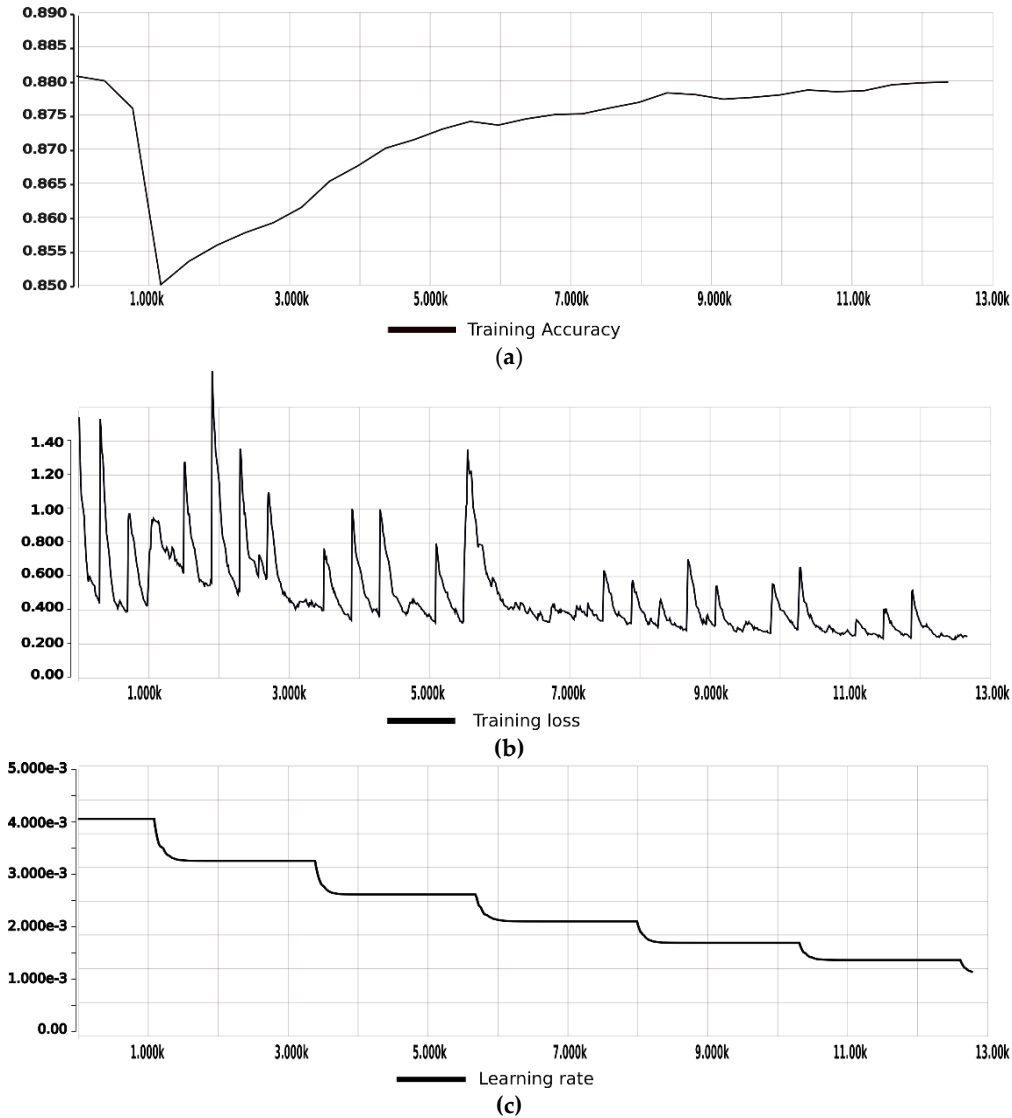

**Figure 7**. Training summaries of A-XCRF, includes **(a)** Training accuracy, **(b)** Training loss, and **(c)** Learning rate.

Fig. 7 shows the training summaries of A-XCRF, including (a) the loss values, (b) the validation accuracy, and (c) the learning rate. The spikes in the training loss occur after every training epoch on the unseen data. The loss values were not decreasing monotoneously, but display a decreasing trend. One notable phenomenon is that after the first learning rate decay, the validation accuracies demonstrate a relatively consistent improvement. This indicates that learning rate treatment and initialization strategy is important when training the A-XCRF.

End-to-end training and parameter update could simplify the A-XCRF learning process.

Generative Adversarial Network (GAN) (Goodfellow et al. 2014) train both of GAN's losses using minimax game theory (Salimans et al. 2016), with finding an equilibrium as the objective function. Using a GAN training style, we can set up both of the A-XCRF loss functions as the same minimax objective function and perform parameter update using the gradient flow and parameter update in the GAN architecture. Although both models are unstable and hard to train, A-XCRF with the minimax game algorithm could potentially be an end-to-end architecture with a better accuracy.

## 5. Conclusions

In this paper, a novel technique for addressing the overfitting issue for automatic classification of pointcloud data is presented. The main contribution of our research is the proposal of an XCRF training algorithm and an A-XCRF layer for training, utilizing the unlabeled part of a dataset to improve model accuracy. To address the overfitting behavior of DNN based models, we introduced a method that is not only capable of using similarity values between features of interest, but also induces a controlled noise in the validated model. In the preprocessing procedures of our work, we sliced the pointcloud data using voxel blocks and used HaG and Intensity as the features of interest. PointCNN was used as a classifier. PointCNN is a deep learning architecture that uses the X-transformed technique to consume irregular and unordered data points using a convolution-like operator. For the post-processing, A-XCRF was used, which can be viewed as a deep learning layer that forces the DNN models to respect the similarity penalties given by the unseen data. A-XCRF is a stack of XCRF modules that penalizes nearby similar points that have the dissimilar predicted label with a strict penalty procedure using a hollow compatibility matrix.

Experiments were carried out using the ISPRS 3D labeling benchmark dataset. Comparisons were made with PointCNN and PointCNN with CRF-RNN. In addition, a comparison with other techniques that has been tested on the benchmark dataset was presented. Experimental results show that our proposed technique was better than the other proposals in term of average F1 Score (71.05%) and the overall accuracy was on par with the current best proposal.

Our research provides a stepping stone for improving the quality of machine learning models by introducing artificial labels and a training procedure for the unseen data by invoking XCRF similarity penalties. Further improvement may be achieved by introducing geometrical features in the XCRF algorithm to better handle classes with a linear geometrical shape and doing end-to-end training of A-XCRF layer using a GAN style architecture.

## Acknowledgement

We would like to thank Ivar Oveland for providing the HaG features. We also would like to thank all the Github authors, whose code we used for the experiments.

## References

Abadi, M., Agarwal, A., Barham, P., Brevdo, E., Chen, Z., Citro, C., Corrado, G. S., Davis, A., Dean, J. & Devin, M. (2016). Tensorflow: Large-scale machine learning on heterogeneous distributed systems. *arXiv preprint arXiv:1603.04467*.

Adams, A., Baek, J. & Davis, M. A. (2010). *Fast high-dimensional filtering using the permutohedral lattice*. Computer Graphics Forum: Wiley Online Library. 753-762 pp.

Arief, H., Strand, G.-H., Tveite, H. & Indahl, U. (2018). Land Cover Segmentation of Airborne LiDAR Data Using Stochastic Atrous Network. *Remote Sensing*, 10 (6): 973.

Armeni, I., Sener, O., Zamir, A. R., Jiang, H., Brilakis, I., Fischer, M. & Savarese, S. (2016). *3d semantic parsing of large-scale indoor spaces*. Proceedings of the IEEE Conference on Computer Vision and Pattern Recognition. 1534-1543 pp.

Atkinson, P. M. & Tatnall, A. (1997). Introduction neural networks in remote sensing. *International Journal of remote sensing*, 18 (4): 699-709.

Blum, A. & Rivest, R. L. (1989). *Training a 3-node neural network is NP-complete*. Advances in neural information processing systems. 494-501 pp.

Chen, L.-C., Papandreou, G., Kokkinos, I., Murphy, K. & Yuille, A. L. (2016). Deeplab: Semantic image segmentation with deep convolutional nets, atrous convolution, and fully connected crfs. *arXiv preprint arXiv:1606.00915*.

Cramer, M. (2010). The DGPF-test on digital airborne camera evaluation–overview and test design. *Photogrammetrie-Fernerkundung-Geoinformation*, 2010 (2): 73-82.

Dahl, G. E., Sainath, T. N. & Hinton, G. E. (2013). *Improving deep neural networks for LVCSR using rectified linear units and dropout*. Acoustics, Speech and Signal Processing (ICASSP), 2013 IEEE International Conference on: IEEE. 8609-8613 pp.

Dai, A., Chang, A. X., Savva, M., Halber, M., Funkhouser, T. A. & Nießner, M. (2017). *ScanNet: Richly-Annotated 3D Reconstructions of Indoor Scenes*. CVPR. 10 pp.

De Boer, P.-T., Kroese, D. P., Mannor, S. & Rubinstein, R. Y. (2005). A tutorial on the cross-entropy method. *Annals of operations research*, 134 (1): 19-67.

Goodfellow, I., Pouget-Abadie, J., Mirza, M., Xu, B., Warde-Farley, D., Ozair, S., Courville, A. & Bengio, Y. (2014). *Generative adversarial nets*. Advances in neural information processing systems. 2672-2680 pp.

Horvat, D., Žalik, B. & Mongus, D. (2016). Context-dependent detection of non-linearly distributed points for vegetation classification in airborne LiDAR. *ISPRS Journal of Photogrammetry and Remote Sensing*, 116: 1-14.

Ioffe, S. & Szegedy, C. (2015). Batch normalization: Accelerating deep network training by reducing internal covariate shift. *arXiv preprint arXiv:1502.03167*.

Krizhevsky, A., Sutskever, I. & Hinton, G. E. (2012). *Imagenet classification with deep convolutional neural networks*. Advances in neural information processing systems. 1097-1105 pp.

Krähenbühl, P. & Koltun, V. (2011). *Efficient inference in fully connected crfs with gaussian edge potentials*. Advances in neural information processing systems. 109-117 pp.

Lafferty, J., McCallum, A. & Pereira, F. C. (2001). Conditional random fields: Probabilistic models for segmenting and labeling sequence data.

LeCun, Y., Jackel, L., Bottou, L., Cortes, C., Denker, J. S., Drucker, H., Guyon, I., Muller, U. A., Sackinger, E. & Simard, P. (1995). Learning algorithms for classification: A comparison on handwritten digit recognition. *Neural networks: the statistical mechanics perspective*, 261: 276.

LeCun, Y., Bengio, Y. & Hinton, G. (2015). Deep learning. *nature*, 521 (7553): 436.

Li, Y., Bu, R., Sun, M., Wu, W., Di, X. & Chen, B. (2018). *PointCNN: Convolution On X-Transformed Points*. Advances in Neural Information Processing Systems. 826-836 pp.

Mallet, C., Bretar, F. & Soergel, U. (2008). Analysis of full-waveform lidar data for classification of urban areas. *Photogrammetrie Fernerkundung Geoinformation*, 5: 337-349.

Maturana, D. & Scherer, S. (2015). *Voxnet: A 3d convolutional neural network for real-time *object recognition*. Intelligent Robots and Systems (IROS), 2015 IEEE/RSJ International Conference on: IEEE. 922-928 pp.

Mikolov, T., Karafiát, M., Burget, L., Černocký, J. & Khudanpur, S. (2010). *Recurrent neural network based language model*. Eleventh Annual Conference of the International Speech Communication Association.

Minsky, M. & Papert, S. A. (1987). *Perceptrons: expanded edition*: MIT press.

Niemeyer, J., Rottensteiner, F. & Soergel, U. (2014). Contextual classification of lidar data and building object detection in urban areas. *ISPRS journal of photogrammetry and remote sensing*, 87: 152-165.

Niemeyer, J., Rottensteiner, F., Sörgel, U. & Heipke, C. (2016). Hierarchical higher order crf for the classification of airborne lidar point clouds in urban areas. *International Archives of the Photogrammetry, Remote Sensing and Spatial Information Sciences-ISPRS Archives 41 (2016)*, 41: 655-662.

Noh, H., Hong, S. & Han, B. (2015). *Learning deconvolution network for semantic segmentation*. Proceedings of the IEEE international conference on computer vision. 1520-1528 pp.

Paris, S. & Durand, F. (2006). *A fast approximation of the bilateral filter using a signal processing approach*. European conference on computer vision: Springer. 568-580 pp.

Podobnikar, T. & Vrečko, A. (2012). Digital Elevation Model from the Best Results of Different Filtering of a L i DAR Point Cloud. *Transactions in GIS*, 16 (5): 603-617.

Ramiya, A. M., Nidamanuri, R. R. & Ramakrishnan, K. (2016). A supervoxel-based spectro-spatial approach for 3D urban point cloud labelling. *International Journal of Remote Sensing*, 37 (17): 4172-4200.

Recht, B., Re, C., Wright, S. & Niu, F. (2011). *Hogwild: A lock-free approach to parallelizing stochastic gradient descent*. Advances in neural information processing systems. 693-701 pp.

Ronneberger, O., Fischer, P. & Brox, T. (2015). *U-net: Convolutional networks for biomedical image segmentation*. International Conference on Medical image computing and computer-assisted intervention: Springer. 234-241 pp.

Salimans, T., Goodfellow, I., Zaremba, W., Cheung, V., Radford, A. & Chen, X. (2016). *Improved techniques for training gans*. Advances in Neural Information Processing Systems. 2234-2242 pp.

Schmidhuber, J. (2015). Deep learning in neural networks: An overview. *Neural networks*, 61: 85-117.

Srivastava, N., Hinton, G., Krizhevsky, A., Sutskever, I. & Salakhutdinov, R. (2014). Dropout: a simple way to prevent neural networks from overfitting. *The Journal of Machine Learning Research*, 15 (1): 1929-1958.

Steinsiek, M., Polewski, P., Yao, W. & Krzystek, P. (2017). Semantische analyse von ALS-und MLS-daten in urbanen gebieten mittels conditional random fields. *Tagungsband der*, 37: 521-531.

Terrasolid. (2016). *TerraScan User's Guide*. Available at: https://www.terrasolid.com/download/tscan.pdf.

Tomasi, C. & Manduchi, R. (1998). *Bilateral filtering for gray and color images*. Computer Vision, 1998. Sixth International Conference on: IEEE. 839-846 pp.

Vosselman, G. & Dijkman, S. (2001). 3D building model reconstruction from point clouds and ground plans. *International archives of photogrammetry remote sensing and spatial information sciences*, 34 (3/W4): 37-44.

Wolf, C. & Jolion, J.-M. (2006). Object count/area graphs for the evaluation of object detection and segmentation algorithms. *International Journal of Document Analysis and Recognition (IJDAR)*, 8 (4): 280-296.

Wu, Z., Song, S., Khosla, A., Yu, F., Zhang, L., Tang, X. & Xiao, J. (2015). *3d shapenets: A deep representation for volumetric shapes*. Proceedings of the IEEE conference on computer

  vision and pattern recognition. 1912-1920 pp.

Yang, Z., Jiang, W., Xu, B., Zhu, Q., Jiang, S. & Huang, W. (2017). A convolutional neural network-based 3D semantic labeling method for ALS point clouds. *Remote Sensing*, 9 (9): 936.

Yang, Z., Tan, B., Pei, H. & Jiang, W. (2018). Segmentation and Multi-Scale Convolutional Neural Network-Based Classification of Airborne Laser Scanner Data. *Sensors*, 18 (10): 3347.

Yousefhussien, M., Kelbe, D. J., Ientilucci, E. J. & Salvaggio, C. (2017). A Fully Convolutional Network for Semantic Labeling of 3D Point Clouds. *arXiv preprint arXiv:1710.01408*.

Zhao, H., Liu, Y., Zhu, X., Zhao, Y. & Zha, H. (2010). *Scene understanding in a large dynamic environment through a laser-based sensing*. Robotics and Automation (ICRA), 2010 IEEE International Conference on: IEEE. 127-133 pp.

Zhao, R., Pang, M. & Wang, J. (2018). Classifying airborne LiDAR point clouds via deep features learned by a multi-scale convolutional neural network. *International Journal of Geographical Information Science*, 32 (5): 960-979.

Zheng, S., Jayasumana, S., Romera-Paredes, B., Vineet, V., Su, Z., Du, D., Huang, C. & Torr, P. H. (2015). *Conditional random fields as recurrent neural networks*. Proceedings of the IEEE international conference on computer vision. 1529-1537 pp.